\documentclass[10pt,twocolumn,letterpaper]{article}
\usepackage{authblk}

\makeatletter
\renewcommand\AB@affilsepx{, \protect\Affilfont}
\makeatother

\usepackage[pagenumbers]{cvpr} 

\usepackage{graphicx}
\usepackage{amsmath}
\usepackage{amssymb}
\usepackage{booktabs}
\usepackage{makecell}
\usepackage[symbol]{footmisc}

\newcommand{\ourmodel}{HIPNet}

\usepackage[pagebackref,breaklinks,colorlinks]{hyperref}

\usepackage[capitalize]{cleveref}
\crefname{section}{Sec.}{Secs.}
\Crefname{section}{Section}{Sections}
\Crefname{table}{Table}{Tables}
\crefname{table}{Tab.}{Tabs.}

\def\cvprPaperID{7708} 
\def\confName{CVPR}
\def\confYear{2022}

\begin{document}

\title{Hierarchical Neural Implicit Pose Network for Animation  and \\ Motion Retargeting}

\author[1,2]{Sourav Biswas \thanks{} }
\author[1]{Kangxue Yin}
\author[1]{Maria Shugrina}
\author[1,3,4]{Sanja Fidler}
\author[1]{Sameh Khamis}

\affil[1]{NVIDIA}
\affil[2]{University of Waterloo}
\affil[3]{University of Toronto}
\affil[4]{Vector Institute}

\twocolumn[{%
\renewcommand\twocolumn[1][]{#1}%
\vspace{-3.5em}
\maketitle
\thispagestyle{empty}
\vspace{-2.5em}
\begin{center}
    \centering
    
    \includegraphics[width=0.95\linewidth]{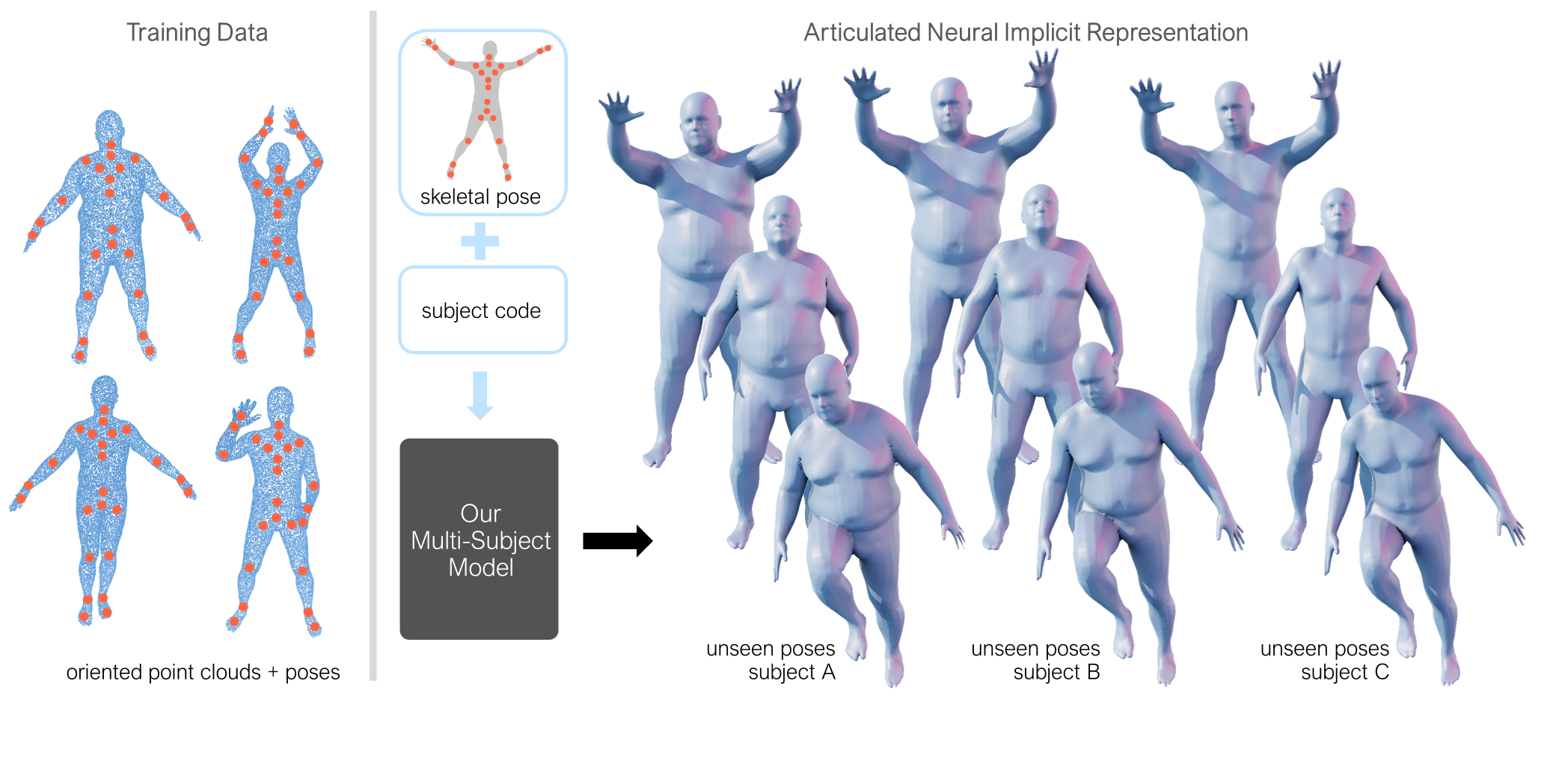}
    \captionof{figure} {Our method models 3D shape of multiple articulated subjects in a single trained neural implicit representation, while showing excellent reconstruction, generalization to novel poses, and only requiring weak supervision. }

    \label{fig:teaser}
\end{center}%
}]

\vspace{0.5ex}

\renewcommand{\thefootnote}{\fnsymbol{footnote}}
\footnotetext[1]{Work done during an internship at NVIDIA.}
\renewcommand*{\thefootnote}{\arabic{footnote}}
\setcounter{footnote}{0}

\begin{abstract}

We present \ourmodel{}, a neural implicit pose network trained on multiple subjects across many poses. \ourmodel{} can disentangle subject-specific details from pose-specific details, effectively enabling us to retarget motion from one subject to another or to animate between keyframes through latent space interpolation. To this end, we employ a hierarchical skeleton-based representation to learn a signed distance function on a canonical unposed space. This joint-based decomposition enables us to represent subtle details that are local to the space around the body joint. 
Unlike previous neural implicit method that requires ground-truth SDF for training, our model we only need a posed skeleton and the point cloud for training, and we have no dependency on a traditional parametric model or traditional skinning approaches. We achieve state-of-the-art results on various single-subject and multi-subject benchmarks.

\end{abstract}

\vspace{-2mm}
\section{Introduction}

Placing controllable characters into a digital 3D world is a long-standing research problem with applications in digital simulation, cinematography, games, AR/VR, and recent metaverse efforts. Traditionally, digital characters are represented as 3D meshes with meticulously crafted control mechanisms (or rigs) for deforming the shape. Explicit control of deformation, for example through the simple but popular linear-blend skinning (LBS) model\cite{lindholm2001user}, inevitably results in artifacts, which countless prior works have attempted to remedy~\cite{liu2014skinning,wang2015linear,jacobson2011bounded}. While it is possible to achieve stunning motion with these traditional rigs, they are also notoriously difficult and time-consuming to design even by skilled 3D artists. 

Recently, the research community has gained interest in implicit or coordinate-based definitions of 3D shapes, for example through learned signed distance functions (SDFs) \cite{park2019deepsdf}. These representations naturally allow capturing intricate, context-dependent variations of geometry, but their application to control and animation of 3D shapes is still in their early stages. Our work follows the promising direction of developing controllable implicit representations for articulated 3D shapes, and makes strides in generalizability, accuracy, and amount of supervision.

We propose to learn a hierarchical part-based signed distance function, conditioned on joint positions and angles as well as a subject identity code. Our learned model, dubbed Hierarchical Implicit Pose Network (or \ourmodel{}), is composed of $n$ \emph{joint-level} SDF sub-networks, arranged hierarchically based on a skeletal graph, and a single \emph{aggregation} network that fuses multiple predictions into the final SDF value.  At run time, a single trained \ourmodel{} network can be used to drive shape changes of multiple characters using skeletal poses unseen in training. We report state-of-the-art results on various multi-subject datasets, showcasing our generalization across subjects and motions. 


While some prior works also use a hierarchical formulation \cite{alldieck2021imghum,deng2020nasa}, crucially ours is the first neural implicit model to train on only point clouds and posed skeletons, without relying on a traditional parametric model  or existing skinning weights. For example, NASA \cite{deng2020nasa} relies on skinning weights of an existing 3D rig in order to learn an implicit representation. As stated earlier, good quality rigs are difficult to obtain, and the promise of implicit functions is precisely to obviate the need for such rigs to begin with. On the other hand, ImGHUM~\cite{alldieck2021imghum} relies on GHUM~\cite{xu2020ghum}, a statistical generative 3D body model with shape and pose spaces. Our work makes an important stride in this direction, enabling training from unlabeled point clouds and driving skeletons that could be the result of 3D capture and pose estimation without heavy post-processing. In addition, our approach disentangles pose and identity, and allows retargeting motion from one character to another, or animating a character by interpolating in the learnt pose space, as shown in Fig~\ref{fig:teaser}. Finally, while we experiment on widely available human datasets, our method is general and can extend to any creatures driven by a known skeleton and 3D pose.

In summary, we propose a pose-driven hierarchical neural SDF model for articulated characters that:
\begin{itemize}
    \item Requires only weak supervision given poses and point clouds, no template meshes or skinning weights.
    \item Disentangles identity from pose without relying on a traditional parametric model, allowing motion retargeting and smooth interpolation between key poses.
    \item Supports multi-subject training and also outperforms single-subject models on all benchmarks.
    \item Is formulated as an SDF, not occupancy, resulting in faster rendering and meshing.
    \item Results in a generative model that generalizes to new subjects via latent code optimization.
\end{itemize}




\section{Related Work}

\newcommand{\YS}{\textbf{Y}}
\newcommand{\NO}{N}
\begin{table*}[h]
\centering
\footnotesize{
\setlength{\tabcolsep}{5pt}
\begin{tabular}{|l|c|c|c|c|c|c|c|} 
\Xhline{1.5pt}
    &  Training & Test-time & Network & Multi-subj. & Extends to & Generative & Generative \\
    &  data & input & output & support & non-humans & shape & pose \\
    \Xhline{1.5pt}
    NASA  \cite{deng2020nasa} & \makecell{occ., \\ skinning weights, poses} & pose & occ. & \NO & \YS & \NO & \YS \\
    \hline
    SNARF  \cite{chen2021snarf} & \makecell{occ., poses \\ \; } & pose & occ. & \NO & \YS & \NO & \YS \\
    \hline
    imGHUM  \cite{alldieck2021imghum} & \makecell{oriented pt.\ clouds, \\ off-srf. pts., \scriptsize{GHUM} enc.} & \scriptsize{GHUM} enc. & \makecell{SDF, \\semantics} & \YS & \NO & \YS${}^*$  & \YS \\
    \hline
    LEAP  \cite{mihajlovic2021leap} & \makecell{occ., verts., \\ skinning weights, poses} & \makecell{shape enc., \\ pose} & occ. & \YS & N & \YS${}^*$ & \YS \\
    \hline
    NPMs  \cite{palafox2021npms} & \makecell{T-pose and posed meshes, \\ dense per-subj. corr.}  & \makecell{learned ID \& pose \\ enc. (not posable)} & SDF & \YS & \YS & \YS & \YS \\
    \hline
    \textbf{Ours}  & oriented pt. clouds, poses & \makecell{learned ID enc., \\ pose} & SDF & \YS & \YS & \YS & \YS \\
 \Xhline{1.5pt}
\end{tabular}
}

\scriptsize{\textit{Legend}: occ.\ - occupancy, pt.\ - point, verts.\ - vertices of the surface mesh, enc.\ - encoding, subj.\ - subject, corr.\ - correspondences, off-surf.\ pts.\ - off-surface points.}
\caption{\textbf{Comparison of Neural Implicit Methods:} Our method requires less supervision and has on-par or better generalization
than existing methods. ${}^*$ imGHUM~\cite{alldieck2021imghum} regresses the pretrained GHUM~\cite{xu2020ghum} shape space and LEAP~\cite{mihajlovic2021leap} regresses the SMPL~\cite{loper2015smpl} shape space, which potentially limits their ability to learn from out-of-distribution observations.
}
\label{table:Methodcomparison}
\end{table*}

\subsection{Classical Approaches}
Constructing poseable 3D character models from observations is a long-standing problem in computer graphics. A number of representations are available for driving the deformation of a 3D shape. For articulated character animation, linear blend skinning (LBS)~\cite{lindholm2001user} is a widely popular approach. In LBS, vertices of a 3D mesh are assigned to the bones and joints of a controlling "skeleton" using vertex-level skinning weights, resulting in a 3D "rig". These weights
are used to propagate bone transformations to the underlying surface geometry. There is a wealth of work on automatically creating LBS-style rigs for a character mesh \cite{baran2007jp,jacobson2011bounded,kavan2010co,le2014ba,RigNet}, however this representation is far from perfect. Simple LBS formula is known to introduce artifacts and sometimes wildly unrealistic deformations (known as the "candy-wrapper" effects), prompting research on alternative formulations ~\cite{kavan2008tv}, and on incorporating pose-specific examples to correct the final shape \cite{lewis2000pose}. Such pose-specific guidance is still used today, and most production rigs are intricate works requiring hours of highly skilled labor. The need for pose-specific control over deformation and the labour-intensive nature of explicit 3D rigs makes neural implicit representations for articulated characters a promising direction for new solutions in this space.

\subsection{Neural Implicit Approaches}
Implicit representations, such as Signed Distance Fields (SDF), have found use in representing and tracking deforming shapes~\cite{curless1996volumetric,newcombe2011kinectfusion}. Most of these grid-based representations faced a trade-off between quality of representation and a prohibitive growth in memory requirements. Various approaches resorted to space-partitioning data structures or hashing functions to address this limitation~\cite{dou2016fusion4d, niessner2013real}. Neural implicit representations~\cite{bhatnagar2020combining,park2019deepsdf,mescheder2019occupancy,chen2019learning,chibane20ifnet} have surpassed these approaches in terms of both quality and scalability.

Within animation, the promise of neural implicit representations is representing rich pose-dependent geometry of
an articulated character \emph{by learning from observations alone}. Our work follows this motivating principle and requires only points and normals sampled from the surface of the deforming geometry and the corresponding skeletal poses for training. While several recent works also aim to represent articulated 3D humans using a deep implicit representation (See Tab.~\ref{table:Methodcomparison}), almost none follow the same low-supervision regime as our method. NASA \cite{deng2020nasa} and imGHUM \cite{alldieck2021imghum} both devise a hierarchical or part-based models, similar to ours, but are supervised by difficult to obtain data: imGHUM requires a parametric template mesh and is trained using existing GHUM parameterization \cite{xu2020ghum}, and NASA training relies on skinning weights, which are not only expensive to generate but might also constrain the model to the space of LBS rigs. Like NASA, LEAP \cite{mihajlovic2021leap} relies on skinning weights, and NPMs \cite{palafox2021npms} need a canonical pose and dense correspondences for every subject. NPMs are also not directly posable, requiring test-time optimization to find a pose encoding for any new pose. In addition, imGHUM \cite{alldieck2021imghum} and LEAP \cite{mihajlovic2021leap}  are closely tied to existing parametric models of the human body, namely GHUM \cite{xu2020ghum} and SMPL \cite{loper2015smpl}, respectively, which makes them difficult to extend to other classes of articulated characters. Additionally, LEAP~\cite{mihajlovic2021leap} and more recent work like SCANimate~\cite{saito2021scanimate} rely on LBS within their model, inheriting the artifacts of traditional skinning. This was also noted in the follow-up work Neural-GIF~\cite{tiwari2021neural}. Out of prior work focused on articulated neural implicit models (Table \ref{table:Methodcomparison}), only SNARF \cite{chen2021snarf} follows our philosophy of low-supervision regime.

While SNARF \cite{chen2021snarf} demonstrates superior results compared to NASA \cite{deng2020nasa}, especially on out-of-domain poses, and is trained in a low-supervision regime, it introduces an expensive root-finding step at inference time and cannot handle multiple subjects with a single network or generalize across identities. The key insight behind SNARF is learning a single pose-independent forward skinning transform to represent the bulk of the deformation, unlike approaches like \cite{mihajlovic2021leap,jeruzalski2020nilbs}, which learn a pose-specific backward skinning function in order to look up occupancy values in the canonical space. Learning pose-specific transform makes these methods vulnerable to over-fitting to the training poses and introduces discontinuities when the same point in the posed space (e.g.\ touching hands) maps to very disparate locations in the canonical space. However, this key insight is also the cause of SNARF's limitations, such as the root-finding step. In our work, we show that a careful set of design decisions can extend simpler feed-forward hierarchical approaches \cite{deng2020nasa,alldieck2021imghum} to superior generalization to novel poses as well as multi-subject training, all in a low-supervision regime and with an ability to generalize beyond humans, like SNARF. As the result, our work is an important step in this new and exciting line of research.

Other recent works address modeling moving human characters in slightly different regime, for example by learning a controllable neural radiance fields from carefully calibrated multi-camera footage \cite{liu2021neural} or by focusing on a mesh representation \cite{Choi_2020_ECCV_Pose2Mesh}. In addition, an extensive line of research on implicit representations focuses on the problem of reconstructing and completing 3D shapes from partial or 2D inputs \cite{chibane2020implicit,chibane2020implicit2,he2020geo,saito2019pifu,saito2020pifuhd}. We omit a detailed discussion of these works to maintain focus on \emph{articulated} neural implicit representations of 3D shape through occupancy or signed distance functions, the subject of our study.

\section{Approach}

\begin{figure*}[t]
\centering
    \vspace{1ex}
  \includegraphics[width=0.98\linewidth]{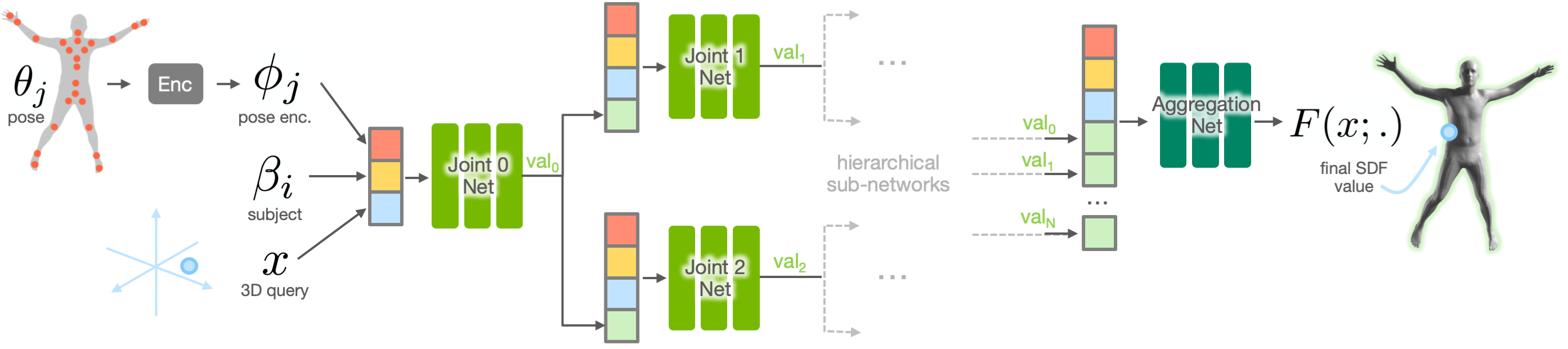}%
\caption{\textbf{Network Overview:} \ourmodel{} is composed
of separate sub-network MLPs arranged hierarchically according to an input skeleton. The final aggregation network is fed the results of these sub-networks and outputs the final SDF value. The subject latent codes $\beta_i$ are optimized during the optimization, alongside the parameters of the pose encoding network $\Phi$. Refer to the text for additional details.}
\label{fig:networks}
\end{figure*}
    
\subsection{Problem Formulation}
Given a dataset of oriented point clouds (\ie, points and normals) for different subjects that share the same underlying skeleton (\eg, human), as well as the 3D pose of that skeleton for each point cloud, our goal is to train a coordinate-based model, conditioned on pose and subject, that generates a valid SDF with a zero-crossing at the given surface points. In effect, we aim to learn an articulated SDF that can be animated by novel skeletal poses and can represent a number of subjects with a single trained model.

\subsection{Our Model}\label{ssec:ourmodel}

\textbf{Overview:} Given a skeletal pose $\theta$ and subject encoding $\beta$, our model \ourmodel{} predicts SDF values at input query points $x \in R^3$. Like some prior modern approaches \cite{deng2020nasa,alldieck2021imghum}, our model is composed of part-based MLPs. We rely on a consistent skeletal structure of the training dataset and structure our model hierarchically, with a separate MLP sub-network for each of $N$ joints in the skeleton. The output of every sub-network $k$ is a signed distance value $F_k(x,.)$ that goes to zero at the surface with a gradient equal to the surface normal. Unlike \cite{deng2020nasa} and \cite{alldieck2021imghum}, we hierarchically propagate the intermediate predictions based on the skeleton's connectivity graph. The output of a sub-network is therefore concatenated with the input and fed to its children down the graph, allowing child nodes more information to refine the region of overlap with the parent. In addition, we train a final aggregation network to fuse the separate predictions of all sub-networks into the final SDF value $F(x,.)$. The overall architecture of our system is illustrated in Fig.~\ref{fig:networks}.

\textbf{Pose Conditioning}: We next discuss how our networks are conditioned on pose. First, the input pose parameters $\theta_j$ (joint transformations) from a sequence $\{\theta_j\}$
are encoded into a latent representation $\phi_j$. Our encoder $\Phi$ is trained jointly with the model in an end-to-end fashion. Thus, $\Phi$ learns to project an input pose into a subspace of the plausible poses that the skeleton can deform into. Note that we ensure that pose parameters for all subjects are first transformed into a canonical global frame. Given that all the subjects share the same skeleton, and therefore are of comparable sizes, this has no negative effect on our model's representational power. The entire pose embedding $\phi_j$ is passed to \emph{all} of the sub-networks and the aggregation network as input, thus making all predictions conditioned on the global pose. This can help disambiguate shape for challenging poses with interaction between different parts of the skeletal graph. Unlike NASA's deformable model~\cite{deng2020nasa}, we employ the hierarchy of the skeleton for propagation and we do not try to learn to cluster the subject into deformable components.

\textbf{Subject Conditioning:} For multi-subject training, our networks are further conditioned on a learned subject embedding $\beta_i \in \mathbb{R}^d$. The subject codes $\beta_i$ are directly optimized during training. This follows the generative latent optimization (GLO) framework~\cite{bojanowski2017optimizing}, where the model is effectively an autodecoder~\cite{park2019deepsdf}. Similarly to the pose embedding, $\beta_i$ conditions all of the sub-networks and the aggregation network. Note that in order to evaluate on datasets with only one subject and to fairly compare to single-subject baselines, we dropped the subject codes from all the inputs to all our networks.

\subsection{Training Data}

To train \ourmodel{}, our method requires oriented 3D point clouds along with subject id and 3D pose parameters of the skeleton for each point cloud. We denote the pose parameters transformed to the canonical global frame with $\theta_{ij}$ for subject $i$ in frame $j$, and their encoding as $\phi_{ij}$. 

\begin{figure}[t]
\vspace{1ex}

\centering
  \includegraphics[width=0.8\linewidth]{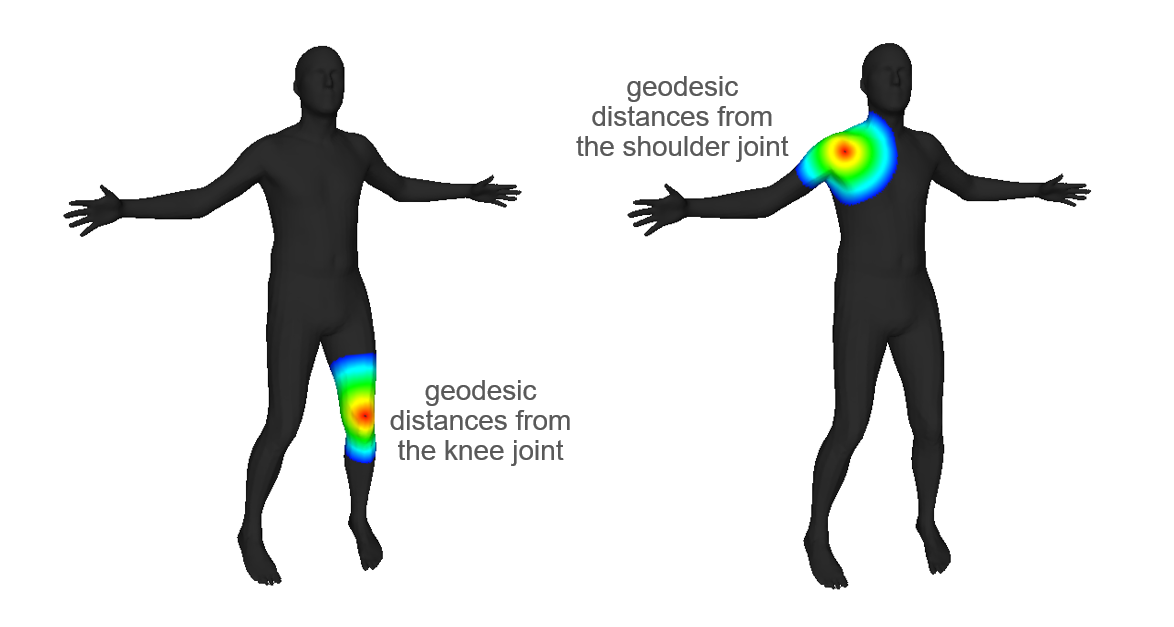}%
\vspace{-3mm}
\caption{We use geodesic distances, pictured, to assign training surface points to the closest joint. The points closest to the joint are used in an additional loss terms to specialize the sub-network corresponding to that joint. This point-joint assignment is not needed during evaluation.}
\label{fig:geodesics}

\end{figure}

We additionally process the input point clouds offline for approximate vertex-joint assignments. To this end, we process each point cloud individually, finding a proxy surface point on for every joint, and then estimating the approximate Geodesic distance between all points and the joints~\cite{crane2013geodesics,liu2012point}. Each point is then assigned to the closest joint. See Figure~\ref{fig:geodesics} for an illustration. This assignment is only needed at training time for an additional loss term and is not needed during evaluation. Please refer to Section~\ref{sec:training} for details.

\subsection{Loss and Training}
\label{sec:training}

\textbf{Weak Supervision:} Unlike prior and concurrent work that requires uniform SDF or occupancy point samples to train an implicit model~\cite{park2019deepsdf,deng2020nasa,palafox2021npms,chen2021snarf}, our method needs point samples only on the \emph{zero-crossing} of the SDF, or the surface of the moving body. Such data is much more readily available, and in principle could be the result of raw capture.  

In a manner similar to IGR~\cite{icml2020_2086}, we utilize geometric initialization~\cite{atzmon2020sal} to initialize the surface to a sphere and train our model to output a solution to the Eikonal equation~\cite{crandall1983viscosity}:

\begin{equation}
    ||\nabla_{x} F(x;.)|| = 1.
\end{equation}

\textbf{Loss Function:} For each input point cloud $\mathcal{P}_{ij}$ with normals $n(x)\ \forall x \in \mathcal{P}_{ij}$, we train our 
model $F:\mathbb{R}^3 \rightarrow \mathbb{R}$ with network parameters $w$ using the following loss function:
\begin{multline} \label{eq:totalloss}
\mathcal{L}(w) = L_\text{surface}(w) + \lambda_\text{eikonal}  L_\text{eikonal}(w) +
\lambda_\text{normal} L_\text{normal}(w) + \\
\sum_{k=1}^{N} \bigg(L^{(k)}_\text{surface}(w) + \lambda_\text{normal} L^{(k)}_\text{normal}(w)\bigg),
\end{multline}

\noindent
where all the $\lambda$ denote hyper-parameters to weigh terms in relative importance.

The data terms in the previous equation are the surface loss $L_\text{surface}$ and the normal loss $L_\text{normal}$, and are defined as:

\begin{equation} \label{eq:surfaceloss}
L_\text{surface}(w) = \sum_{i,j} \sum_{x \in \mathcal{P}_{ij}} |F(x, \phi_{ij}, \beta_i; w)|,
\end{equation}

\begin{equation} \label{eq:normalloss}
L_\text{normal}(w) = \sum_{i,j} \sum_{x \in \mathcal{P}_{ij}} ||\nabla_x F(x, \phi_{ij}, \beta_i; w) - n(x)||,
\end{equation}

\noindent
where $\phi_{ij} = \Phi(\theta_{ij}; w)$ is the encoded pose.

The Eikonal regularization term $L_\text{eikonal}$ is defined as:

\begin{equation} \label{eq:eikonalloss}
L_\textrm{eikonal}(w) = \mathbb{E}(||\nabla_x F(x, .;w) || - 1)^2,
\end{equation}

where we sample points uniformly during training within a canonical bounding box as well as near the surface to enforce the unit-norm constraint on the gradient of the SDF.

\textbf{Sub-network Loss:} The surface and normal losses denoted $L^{(k)}_\text{surface}$ and $L^{(k)}_\text{normal}$, respectively, are added to specialize the $N$ sub-networks to points close to their corresponding joints. The assignment of points to joints is estimated using an approximate geodesic distance on the input point clouds directly~\cite{crane2013geodesics,liu2012point}. In practice, we saw significant improvement over our metrics by encouraging this sub-network specialization. We ablate this choice in our experiments.

Estimating the geodesic distance on the input point clouds is only needed during training. During evaluation we simply feed our input points through the entire network. We observe that the sub-network that tends to have the highest impact on the final SDF value is the one corresponding to the joint closest to the input point.

\subsection{Implementation Details}
In all our experiments, we set $\lambda_\text{eikonal} = 0.1$ and $\lambda_\text{normal} = 0.1$. The surface terms have no associated weight.

The subject codes are randomly initialized from $\mathcal{N}(0, 10^{-4}) \in \mathbb{R}^{256}$, a zero-mean Gaussian with $10^{-4}$ standard deviation, where the dimensionality of the subject latent space is $256$.

All our MLPs are of size $256 \times 8$ with a single skip connection at layer 4, similar to~\cite{park2019deepsdf}. We train on individual points clouds sampled at 25k points per batch. We sample another 25k points uniformly within the bounding box and near the surface for the Eikonal regularization term.

We train \ourmodel{} using ADAM~\cite{kingma2014adam} with a fixed learning rate of $10^{-4}$. Our training runs for 50 epochs on 4 Tesla V100 GPUs for a little over 3 days.

\section{Experiments}

We describe our datasets in section~\ref{sec:datasets}, evaluate our model qualitatively and quantitatively and compare to the state-of-the-art methods in~\ref{sec:evaluation}, and finally conclude by ablating the various design choices in~\ref{sec:ablation}.

\subsection{Datasets}
\label{sec:datasets}

We trained our model on various motion capture datasets. We used various datasets of minimally clothed humans from the AMASS motion capture archive~\cite{mahmood2019amass}, in addition to CAPE~\cite{ma2020cape}, the recent clothed human dataset.
\begin{itemize}
\item \textbf{Transitions} is a single subject dataset with 110 motion sequences. We randomly selected 80 sequences for training and the remaining 30 for testing.
\item \textbf{CMU}\footnote{The homepage of the original dataset is at http://mocap.cs.cmu.edu.} is a dataset of multiple subjects. Initially released as the Human Interaction subset of the CMU Motion Capture dataset, it was re-exported and added to AMASS. CMU contains over 100 subjects, but we dropped subjects with less than 5 motion sequences, leaving 89 subjects for our evaluation.
\item \textbf{DFAUST}~\cite{bogo2017dfaust} is another dataset that has 10 unique subjects performing varying motion sequences.
\item \textbf{CAPE}~\cite{ma2020cape} is a dataset of real scans with 10 male and 5 female subjects in various clothing combinations. While the total number of subjects is 15, there are 41 unique subject-clothing combinations overall.
\end{itemize}

For the multi subject datasets, we select 75\% of sequences for training and 25\% for each subject so no subjects are unseen during inference, similar to recent work~\cite{mihajlovic2021leap,palafox2021npms}. Testing on unseen subjects is explored later in this section. We also randomly sample frames within each motion sequence when dealing with larger datasets. We randomly sampled 3\% of the frames within each sequence for CMU and 50\% of the frames in each sequence for CAPE. Overall with both train and test sets, there are over 80k, 80k, 30k, and 70k samples used in the Transitions, CMU, DFAUST, and CAPE datasets respectively.

For all these experiments we randomly sample 25k surface points alongside their normals on the input meshes. While all of these datasets rely on SMPL~\cite{loper2015smpl} for the consistent topology and the skinning weights, our approach is completely agnostic to both.

\subsection{Evaluation}
\label{sec:evaluation}

\textbf{Metrics:} We report the mean Intersection over Union (mIoU) and the Chamfer-L1 distance~\cite{fan2017point}. The mIoU is calculated using points randomly sampled within a tight bounding box around the subject which is stretched along the diagonal by 10\%~\cite{deng2020nasa}. We also report the mIoU for near surface points for our method which has sampled by taking surface points from the ground truth mesh and then adding isotropic noise with $\mathcal{N}(0.0, 0.03)$, a zero-mean Gaussian with $0.03$ standard deviation.

The Chamfer-L1 distance is the mean of an accuracy and a completeness metric. The accuracy metric is the mean distance of points on the reconstructed mesh to their closest point on the ground truth mesh, and the completeness metric is defined in the same way in the opposite direction. Similar to prior work~\cite{mescheder2019occupancy,fan2017point}, we randomly sample 100k points from both meshes and use $1/10$ times the maximal edge length of the mesh’s bounding box as unit $1$ to normalize our distance estimation.

\textbf{Baselines:}
We compare to the reported mIoU for various recent
baselines, NASA~\cite{deng2020nasa}, NPM~\cite{palafox2021npms}, and SNARF~\cite{chen2021snarf}. We also reproduced NASA within our setup. In that setup, we reduced the number of surface and non-surface point samples to 25k from the original 100k, and we switched the surface point sampling strategy to triangle area-weighted uniform sampling instead of Poisson-Disc sampling.

\textbf{Quantitative Results:}
We first compare our model, \ourmodel{}, to NASA using the same point sampling strategy, the same number of points per mesh, and the same dataset split. We report the uniform mean IoU in Tab.~\ref{table:nasa}. As shown, both our single and multi subject models outperform the deformable NASA model on all four datasets. Notably, NASA significantly underperforms on CAPE where subjects are the most diverse. It is worth noting that NASA requires skinning weights as additional input while our model (both variants) does not need any input beyond the point cloud and the 3D pose.

\begin{table}[t]

\centering
\small{
\setlength{\tabcolsep}{5pt}
\begin{tabular}{@{}|c| c c c c|} 
 \hline
 mIoU $\uparrow$ (\%)  & Transitions & CMU & DFAUST & CAPE \\ [0.5ex] 
 \hline
 Ours (multi subj) & 96.64 & \textbf{97.15} & \textbf{97.39} & 89.27 \\ 
 Ours (single subj) & \textbf{97.74} & 95.55 & 96.24 & \textbf{89.45} \\
 NASA & 95.21 & 89.96 & 92.43 & 75.65 \\[1ex] 
 \hline
\end{tabular}
}
\vspace{-2mm}
\caption{Uniform mean IoU comparison against NASA~\cite{deng2020nasa} with the same setup on all datasets.}
\label{table:nasa}
\vspace{-4mm}
\end{table}

We next report our uniform and near-surface mean IoU on DFAUST and CAPE in Tab.~\ref{table:dfaust} and Tab.~\ref{table:cape}, respectively. With the uniform IoU setting, all methods performed competitively. The surprising exception is NPM~\cite{palafox2021npms}, which wile very competitive on CAPE, underperformed significantly on DFAUST. Our numbers are  state-of-the-art on both datasets and in all metrics, outperforming all baselines.

\begin{table}[t]

\centering
\small{
\setlength{\tabcolsep}{6pt}
\begin{tabular}{@{}|c| c c|} 
 \hline
  & \multicolumn{2}{c|}{DFAUST} \\ [0.5ex] 
  mIoU $\uparrow$ (\%)  & Uniform & Near-Surface \\
 \hline
  Ours (multi subj) & \textbf{97.39} & \textbf{95.61} \\
  Ours (single subj) & 96.24 &  93.85 \\
  SNARF & 97.31 & 90.38\\
  NASA & 96.14 & 86.98 \\
  NPM & 83 & - \\
 \hline
\end{tabular}
}
\vspace{-2mm}
\caption{Uniform and near-surface mean IoU comparison on the DFAUST~\cite{bogo2017dfaust} dataset against SNARF, NASA, and NPM using reported numbers.}
\label{table:dfaust}
\vspace{-2mm}
\end{table}

\begin{table}[t]
\centering
\small{
\setlength{\tabcolsep}{6pt}
\begin{tabular}{@{}|c| c c|} 
 \hline
 & \multicolumn{2}{c|}{CAPE} \\ [0.5ex] 
  mIoU $\uparrow$ (\%) & Uniform & Near-Surface \\
 \hline
  Ours (multi subj) & \textbf{87.70} & \textbf{82.48} \\
  Ours (single subj) &  84.96 & 79.07 \\
  NPM & 83 & - \\
 \hline
\end{tabular}
}
\vspace{-2mm}
\caption{Uniform mean IoU comparison on CAPE~\cite{ma2020cape} against NPM using reported numbers.}
\label{table:cape}
\vspace{-2mm}
\end{table}

\begin{figure*}[t]
\centering
  \includegraphics[width=0.75\linewidth]{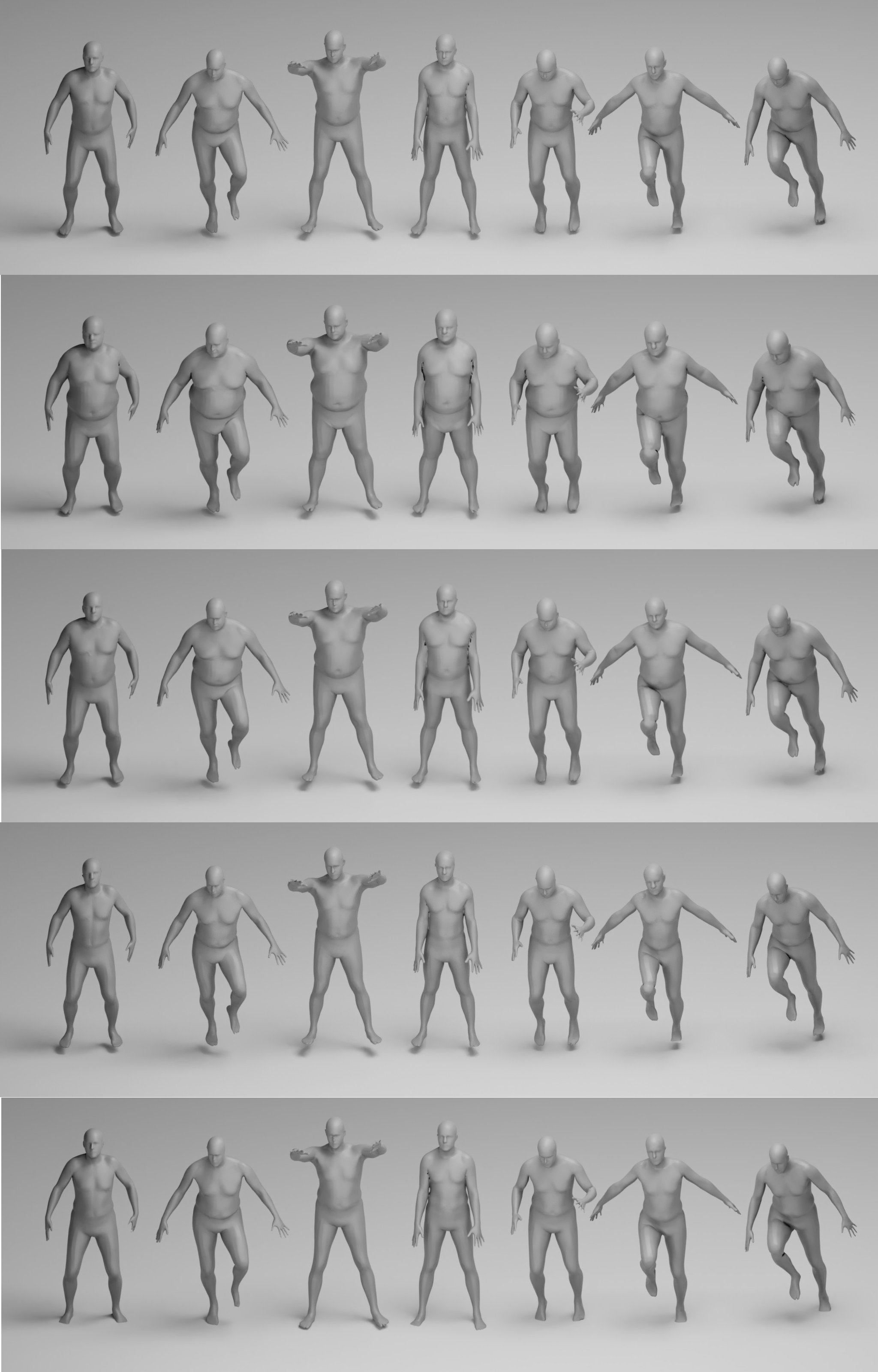}%
  \vspace{-2mm}
\caption{Qualitative results on DFAUST showcasing various unique subjects performing the same test-time poses.}
\label{fig:gallery}
 \vspace{-2mm}
\end{figure*}

\begin{table}[t]
\centering
\small{
\setlength{\tabcolsep}{5pt}
\begin{tabular}{@{}|c| c c c c|} 
 \hline
 Chamfer-L1 $\downarrow$ & Transitions & CMU & DFAUST & CAPE \\ [0.5ex] 
 \hline
 Ours (multi subj) & \textbf{3.625} & \textbf{4.5} & \textbf{1.95} & \textbf{139.5} \\ 
 Ours (single subj) & 3.875 & 6.125 & 4.0 & 181.5 \\
 \hline
\end{tabular}
}
\vspace{-2mm}
\caption{The Chamfer-L1 metric ($\times 10^{-4}$) on all datasets. }
\label{table:chamfer}
\vspace{-2mm}
\end{table}

Finally, we report the Chamfer-L1 distance on all four datasets in Tab.~\ref{table:chamfer}. We report the Chamfer-L1 distance scaled per-mesh such that $1/10$ times the maximal edge length of the mesh’s bounding box as unit $1$~\cite{liu2012point,mescheder2019occupancy}. Our multi-subject model consistently outperforms the single-subject model on this metric.

\textbf{Qualitative Results:} A gallery of our extracted meshes for 5 unique subjects from DFAUST performing the same set of test-time poses is rendered in Fig.~\ref{fig:gallery}. By varying the subject code and maintaining the same pose we effectively perform motion retargeting for in-distribution subjects.

\begin{table}[t]
\vspace{1ex}

\centering
\setlength{\tabcolsep}{2pt}
\begin{tabular}{|p{4.4cm}| c |} 
 \hline
 mIoU $\uparrow$  (\%) & DFAUST \\ [0.5ex] 
 \hline
 base model & 86.55 \\ 
 + sub-network losses & 96.17 \\
 \hspace{2ex} + hierarchical aggregation & 96.24  \\
 \hspace{4ex} + subject latent codes & \textbf{97.39}  \\ [1ex] 
 \hline
\end{tabular}
\vspace{-3mm}
\caption{Ablation study on DFAUST.}
\label{table:ablation}
\vspace{1ex}
\end{table}

\subsection{Ablation Studies}
\label{sec:ablation}

We next performed various ablations on DFAUST to evaluate the impact of various design choices on the reported metrics. We report our findings in Tab.~\ref{table:ablation} to motivate our final network design. The baseline model does not use the sub-network loss and aggregates the sub-network predictions in one step (\ie, a flat hierarchy). The biggest impact is clearly due to the sub-network specialization, significantly contributing to the final performance. The hierarchical aggregation and the latent codes further improve the performance as well.

\section{Discussion}
\subsection{Applications}

\textbf{Animation:} Unlike various prior works, our model does not require a traditional morphable model during training. Our shape space is entirely learnt and is not based on a pre-trained shape space of SMPL~\cite{loper2015smpl} or GHUM~\cite{xu2020ghum}. This makes our model suitable for downstream tasks in animation and motion retargeting, where only the skeleton and 3D pose might be available. This also makes it suitable to animate characters or animals that cannot envelope a human skeleton.

\begin{figure}[t]
  \vspace{-4mm}
\centering
  \includegraphics[width=0.99\linewidth,trim=0 40 40 90,clip]{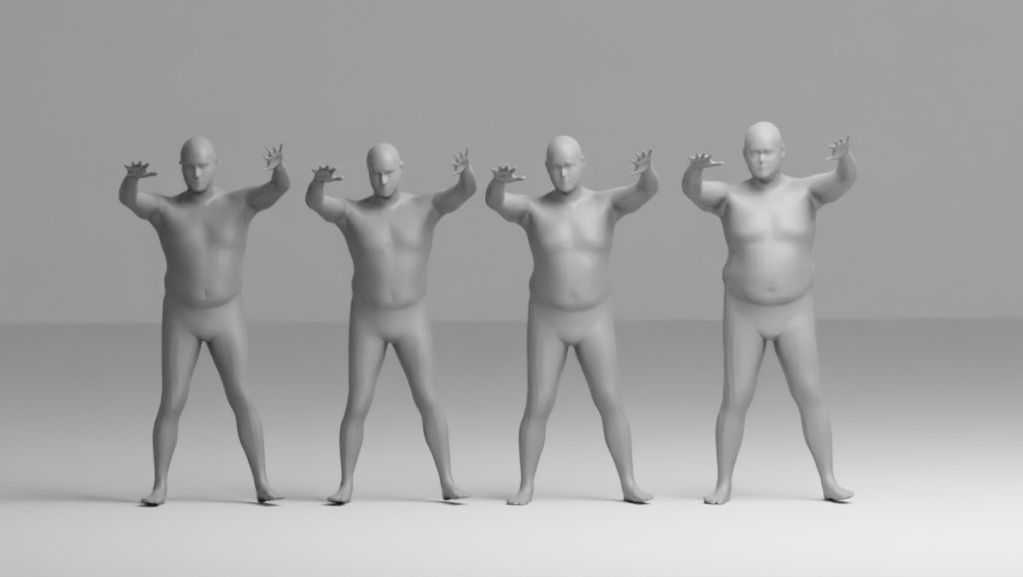}%
  \vspace{-2mm}
\caption{Random subjects sampled from a multivariate Gaussian fitted to our subject codes and rendered in the same pose.}
\label{fig:rand}
\vspace{-1mm}
\end{figure}

\textbf{Generating New Subjects:} Following the Generative Latent Optimization (GLO) framework~\cite{bojanowski2017optimizing}, we can fit a multivariate Gaussian with a full covariance matrix to our subject latent codes. We can then sample from that distribution new characters to showcase the diversity of the data and the representational power of the model. We showcase some samples from our model in Fig.~\ref{fig:rand}.

\begin{figure}[t]
\centering
  \includegraphics[width=0.99\linewidth]{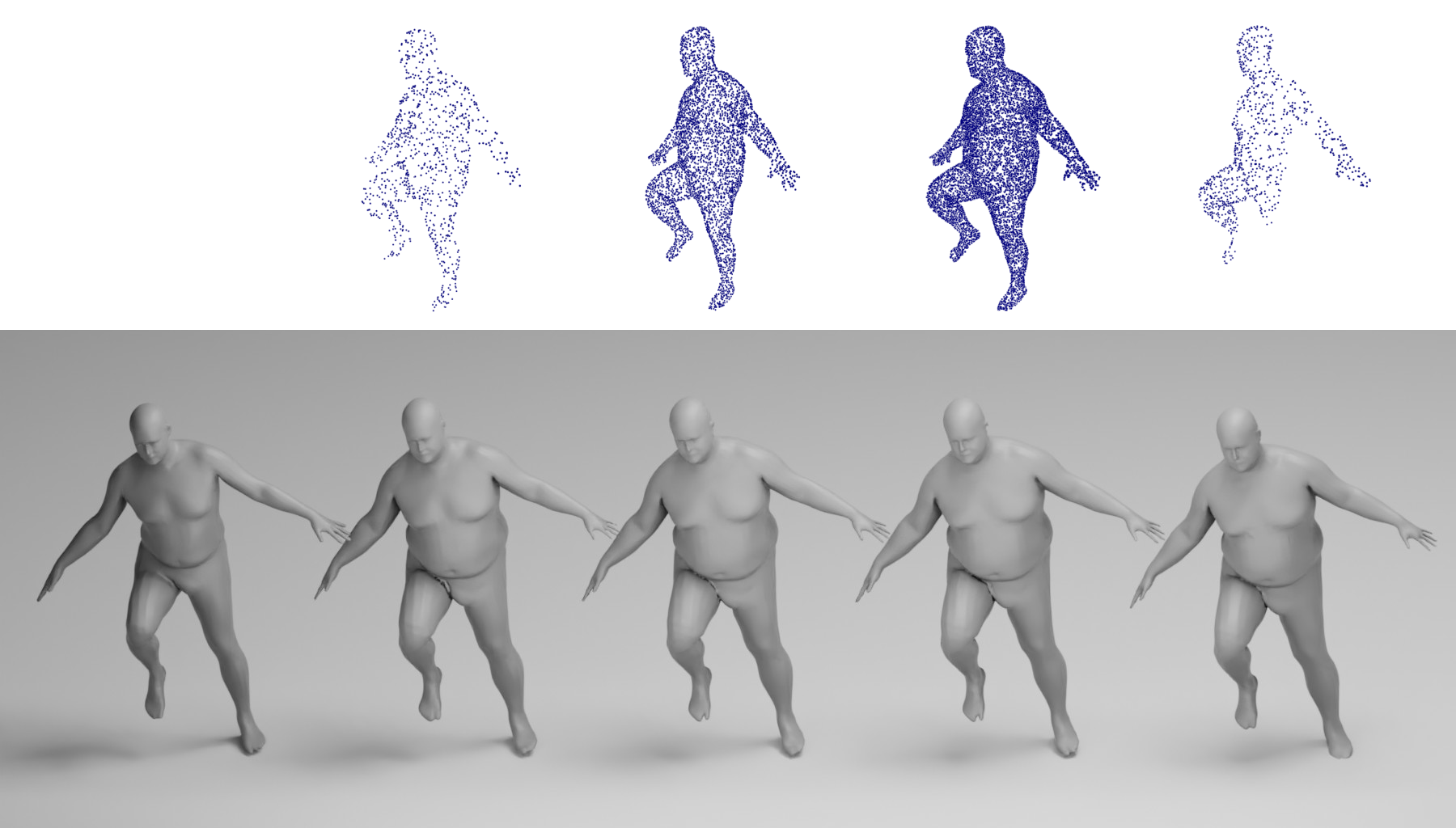}%
  \vspace{-2mm}
\caption{Test-time optimization fitting a point cloud. The number of points each point cloud has (left to right): 0, 1k, 5k, 10k, and 1k (front-only). The mIoU for the reconstructed meshes are (left to right): 68.42\%, 93.82\%, 94.21\%, 97.90\%, and 96.32\%.}
\label{fig:opt}
\vspace{-1ex}
\end{figure}

\textbf{Test Time Optimization:} Similar to~\cite{palafox2021npms}, we can apply test-time optimization to fit a subject code to an input point cloud, even for an out-of-distribution subject. This approach parallels traditional online fitting of shape space parameters, studied in earlier body and hand personalization work~\cite{bogo2016keep,tan2016fits,tkach2017online,shotton2011real}. We showcase this application in Fig.~\ref{fig:opt}. The code is initialized to the mean of the subject codes and optimized for 100 iterations with a learning rate of $10^{-3}$. The mIoU and quality of the predicted mesh increase with the number of input points. More crucially, with only 1k points on the front of the mesh we are competitive with 10k points uniformly sampled everywhere. This specific settings simulates points coming from a front-facing depth sensor and is of practical importance.

\subsection{Limitations}
One major limitation of our work stems from the inability of implicit models to represent open and thin surfaces. This is negatively impacting results on datasets with clothed humans, and generally thin extremities like fingers, tails, and so on. There is a recent interest in investigating the right representation for these surfaces~\cite{chibane2020neural}.

Another question that is generally under-investigated in the field is the impact of modeling muscles on surface deformations. Our work in a way overcomes the limitations of LBS~\cite{lindholm2001user} and PSD~\cite{lewis2000pose} by directly learning to deform from the data. However, modeling the physics of muscles will likely yield a much more data-efficient approach that relies equally on priors from the physical world.
\section{Conclusion}
\label{sec:conc}

In this paper we introduced \ourmodel{}, a hierarchical neural implicit pose network that can represent a variety of subjects under different poses. Unlike prior work, we do not need a traditional morphable model, skinning weights, or dense correspondences during training. Our model can learn directly from input point clouds and 3D poses. Consequently, our model can learn to represent any class of objects, and not just humans, assuming a common skeleton is available. This makes it suitable for downstream animation and motion retargeting tasks. We report state-of-the-art results on four datasets, outperforming recent baselines. In the future we would like to incorporate a render-and-compare pipeline to learn directly from depth images. We also plan to investigate neural implicit representations for clothes, which is an open and interesting problem.

{\small
\bibliographystyle{ieee_fullname}
\bibliography{main}
}

\appendix
\section{Network Architecture}

\begin{figure}[h!]
\centering
    \vspace{1ex}
  \includegraphics[width=0.98\linewidth]{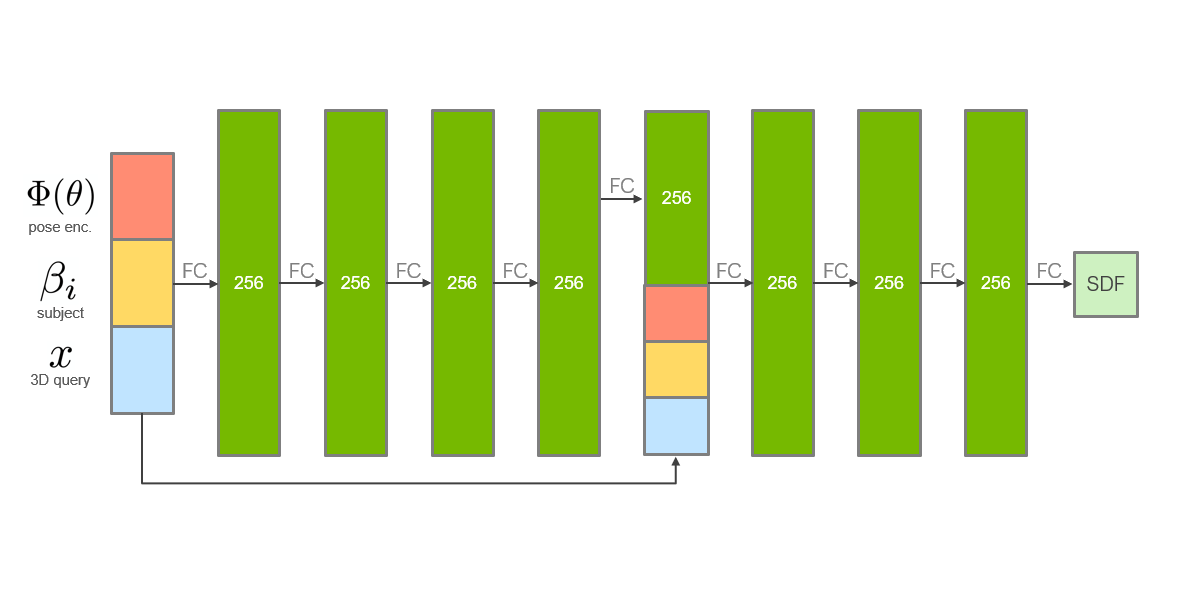}%
\caption{The details of the MLP that is the backbone of our model.}
\vspace{1ex}
\label{fig:mlp}
\end{figure}

Our network is composed of a series of MLPs arranged in a hierarchical skeletal structure. We use 22 MLPs for the 22 joints in the AMASS dataset~\cite{mahmood2019amass}, which is based on the SMPL skeleton but not including the two hand joints~\cite{loper2015smpl}. Each of our MLPs is a series of 8 full-connected layers (of size 256) with a skip connection from the input to layer 4, as shown in Fig.~\ref{fig:mlp}. The input to our MLPs is the encoded pose parameters $\phi = \Phi(\theta)$, the subject code $\beta_i$, and the query point $x$. The subject code is based on the autodecoder framework and it optimized during training~\cite{park2019deepsdf,bojanowski2017optimizing}. For all but the root MLP, the input is also augmented with the output of the parent MLP in the structure.

\section{Additional Results}

We report a full breakdown of our quantitative results on all datasets in Tab.~\ref{table:alliou}. The multi-subject model consistenyl outperforms the single-subject datasets on the multi-subject datasets. On the only single-subject dataset (Transitions), the multi-subject model took significantly longer to train and still was unable to outperform the single-subject model. We expect that eventually it would, but the multi-subject code in this case is a hinderance to the performance.
As expected, the near-surface IoU for the joints of the extremities is lower than that of the less articulated and flexible joints like the pelvis or the neck. This is particularly clear for the wrists because of the higher flexibility of the hands and their greater capacity for articulation. On CAPE~\cite{ma2020cape} for instance, the IoU for the wrists is over $20\%$ lower than that of any other joint. It should be noted here that despite AMASS having a single joint (the wrist) for the entire hand, the hands still articulate fully in the sequences~\cite{mahmood2019amass}. If AMASS is to incorporate a fully-articulated hand model like MANO (which is part of SMPL+H)~\cite{mano_siggraphasia2017}, we expect the IoU for the wrists to improve as well. Hierarchical propagation is in part motivated by propagating predictions from stable, and easy to track, joints to flexible, and hard to track, joints.

\begin{table*}[t]
\centering
\small{
\setlength{\tabcolsep}{5pt}
\begin{tabular}{@{}|l| c c c c| c c c c|} 
 \hline
 & \multicolumn{4}{c|}{Single-Subject} & \multicolumn{4}{c|}{Multi-Subject} \\
 mIoU $\uparrow$ (\%) & Transitions & CMU & DFAUST & CAPE & Transitions & CMU & DFAUST & CAPE \\
 \hline
 Uniform & 97.74 & 95.55 & 96.24 & 84.96 & 96.64 & 97.15 & 97.39 & 87.70 \\
 Near-Surface & 96.55 & 93.20 & 93.85 & 79.07 & 95.00 & 95.49 & 95.61 & 82.48 \\
 \hline
0) Pelvis & 97.93 & 95.76 & 95.17 & 80.48 & 97.85 & 97.55 & 96.92 & 84.45 \\
1) L-Hip & 96.52 & 94.48 & 94.68 & 83.61 & 96.45 & 96.46 & 96.74 & 88.06 \\
2) R-Hip & 93.69 & 91.56 & 93.20 & 83.26 & 93.43 & 94.17 & 95.07 & 87.81 \\
3) Spine1 & 98.36 & 95.98 & 96.22 & 79.19 & 98.36 & 97.81 & 97.84 & 83.08 \\
4) L-Knee & 97.64 & 95.41 & 95.90 & 82.62 & 87.96 & 97.46 & 97.32 & 86.90 \\
5) R-Knee & 98.34 & 95.84 & 95.47 & 83.11 & 97.48 & 97.58 & 97.00 & 87.72 \\
6) Spine-2 & 97.35 & 94.63 & 95.66 & 78.79 & 97.67 & 96.96 & 97.06 & 81.91 \\
7) L-Ankle & 97.13 & 93.46 & 94.20 & 74.94 & 78.15 & 97.09 & 96.25 & 82.33 \\
8) R-Ankle & 97.51 & 94.48 & 93.57 & 75.66 & 97.04 & 97.00 & 95.92 & 83.83 \\
9) Spine3 & 98.03 & 94.55 & 95.29 & 80.62 & 98.01 & 97.34 & 97.20 & 83.68 \\
10) L-Foot & 89.20 & 84.82 & 87.00 & 68.05 & 89.09 & 95.09 & 90.19 & 74.31 \\
11) R-Foot & 92.82 & 89.54 & 84.37 & 68.99 & 94.17 & 95.24 & 89.48 & 73.60 \\
12) Neck & 93.97 & 91.68 & 94.01 & 81.92 & 93.92 & 92.98 & 95.52 & 84.07 \\
13) L-Collar & 97.60 & 94.27 & 95.71 & 82.59 & 97.48 & 96.20 & 96.94 & 84.53 \\
14) R-Collar & 97.58 & 94.20 & 95.34 & 82.82 & 97.37 & 96.47 & 96.71 & 84.41 \\
15) Head & 91.54 & 88.41 & 92.87 & 78.98 & 91.48 & 89.71 & 94.30 & 81.62 \\
16) L-Shoulder & 95.23 & 90.62 & 92.45 & 81.10 & 96.61 & 93.45 & 93.40 & 82.29 \\
17) R-Shoulder & 94.43 & 89.88 & 90.94 & 81.17 & 95.71 & 93.60 & 91.97 & 82.94 \\
18) L-Elbow & 97.18 & 90.74 & 91.15 & 73.63 & 93.76 & 92.58 & 92.67 & 73.51 \\
19) R-Elbow & 96.43 & 90.99 & 89.71 & 73.45 & 96.41 & 90.91 & 91.63 & 75.22 \\
20) L-Wrist & 91.71 & 84.54 & 85.86 & 49.95 & 90.99 & 89.75 & 90.06 & 49.55 \\
21) R-Wrist & 91.37 & 84.63 & 86.53 & 49.78 & 90.95 & 89.79 & 90.10 & 50.54 \\
 \hline
\end{tabular}
}
\vspace{-2mm}
\caption{A breakdown of the mean IoU for our single-subject and multi-subject model on all datasets. }
\label{table:alliou}
\vspace{-2mm}
\end{table*}

\section{Ethical Considerations}

\subsection{Social Impact}

It is worth noting that some datasets we used are known to have an imbalance across gender and ethnicity. CAPE~\cite{ma2020cape} for instance has 10 male subjects to 5 female subjects. Generally speaking, collecting a dataset of full human bodies that spans genders, ethnicities, and body shapes is a very challenging task. This is in part due to local and international regulations around the collection and use of human data, as well as due to cultural and social barriers.

Our model has applications in full body tracking with a personalized body model. This has various downstream applications in digital simulation, cinematography, games, AR/VR, and recent metaverse efforts~\cite{shotton2011real,tan2016fits,dou2016fusion4d}. Tracking people for any purpose is a sensitive topic that can potentially violate their right to privacy. Downstream applications of our model should present end-users with an informed agreement that clearly details the use of their data and allow them the right to retract their consent at any point in the future.

\subsection{Human Data}
The AMASS and CAPE datasets were collected through the Max Planck Institute for Intelligent Systems (MPI-IS)~\cite{mahmood2019amass,ma2020cape}. According to the authors, all subjects were given prior, written, informed consent for the capture and use of their data for research purposes. Additionally, the experimental procedure and consent form were reviewed by the University of Tuebingen Ethics Committee with no objections.

\end{document}